\documentclass[USenglish]{article}

\ifx\directlua\undefined\ifx\XeTeXcharclass\undefined
  \usepackage[utf8]{inputenc}                           
\usepackage[big]{dgruyter}

\usepackage{enumitem}               

\newlist{tabitemize}{itemize}{1}    
\setlist[tabitemize]{label=\textbullet, nosep, leftmargin=*}
                     
\usepackage{subcaption}             
\usepackage{tabularx}               
\newcolumntype{Y}{>{\centering\arraybackslash}X}    
\newcolumntype{Z}{>{\hspace{1em}}X}
\renewcommand\tabularxcolumn[1]{m{#1}}              

\usepackage{pgfplots}               
\pgfplotsset{compat=1.18}
\usepackage{multicol}               
\usepackage{siunitx}

\usepackage[backend=biber]{biblatex}
\usepackage[acronym]{glossaries}
\makeglossaries
\newacronym{fmi}{FMI}{Functional Mock-up Interface}
\newacronym{fmu}{FMU}{Functional Mock-up Unit}
\newacronym{osi}{OSI}{Open Simulation Interface}
\newacronym{osmp}{OSMP}{OSI Sensor Model Packaging}
\newacronym{ros}{ROS}{Robot Operating System}
\newacronym{devs}{DEVS}{Discrete Event System Specification}
\newacronym{thw}{THW}{time headway}

\begin{document}

    \title{Integration of an Agent Model into an Open Simulation Architecture for Scenario-Based Testing of Automated Vehicles}
    
    \articletype{Applications}
    
    \author*[1]{{Christian} {Geller}}
    
    \author[2]{{Daniel} {Becker}}
    
    \author[1]{{Jobst} {Beckmann}}
    
    \author[1]{{Lutz} {Eckstein}}
    
    \affil[1]{{Institute for Automotive Engineering, RWTH Aachen University - Vehicle Intelligence \& Automated Driving}}
    
    \affil[2]{{The author has left Institute for Automotive Engineering, RWTH Aachen University, but all contributions are related to his former employment.}}

    \runningauthor{Christian Geller et al.}
    \runningtitle{Integration of an Agent Models into an Open Simulation Architecture}
    
    \abstract{
    Simulative and scenario-based testing are crucial methods in the safety assurance for automated driving systems. To ensure that simulation results are reliable, the real world must be modeled with sufficient fidelity, including not only the static environment but also the surrounding traffic of a vehicle under test. Thus, the availability of traffic agent models is of common interest to model naturalistic and parameterizable behavior, similar to human drivers. The interchangeability of agent models across different simulation environments represents a major challenge and necessitates harmonization and standardization.
    To address this challenge, we present a standardized and modular simulation integration architecture that enables the tool-independent integration of traffic agent models. The architecture builds upon the Open Simulation Interface (OSI) as a structured message format and the Functional Mock-up Interface (FMI) for dynamic model exchange. 
    Rather than introducing yet another model or simulation tool, we provide a reusable reference implementation that translates these standards into a practical integration blueprint, including clear interfaces, data mappings, and execution semantics.
    The generic nature of the architecture is demonstrated by integrating an exemplary agent model into three widely used simulation environments: OpenPASS, CARLA, and CarMaker. As part of the evaluation, we show that the model yields consistent behavior in all simulation platforms, thereby validating the interoperability, modularity, and standard compliance of the proposed architecture. The reference implementation lowers integration barriers, serves as a foundation for future research, and is made publicly available at \href{https://github.com/ika-rwth-aachen/agent-model-integration}{github.com/ika-rwth-aachen/agent-model-integration}.
    }
        

    \transabstract[german]{
    Simulative und szenarienbasierte Tests sind zentrale Methoden zur Absicherung automatisierter Fahrsysteme. Damit Simulationsergebnisse belastbar sind, muss die reale Welt mit ausreichender Modellgüte abgebildet werden, einschließlich des umgebenden Verkehrs eines automatisierten Fahrzeugs. Dafür werden Agentenmodelle benötigt, die sowohl naturalistisches als auch parametrisierbares Verhalten realistisch modellieren können. Eine wesentliche Herausforderung besteht dabei in der Integration und Austauschbarkeit von Agentenmodellen über unterschiedliche Simulationsumgebungen hinweg und erfordert eine weitergehende Harmonisierung und Standardisierung.
    Um dieser Herausforderung zu begegnen, stellen wir eine standardisierte und modulare Integrationsarchitektur vor, die die simulationsunabhängige Einbindung von Verkehrsagentenmodellen ermöglicht. Die Architektur basiert auf dem Open Simulation Interface (OSI) als strukturiertem Nachrichtenformat sowie dem Functional Mock-up Interface (FMI) für den dynamischen Modellaustausch. Die Allgemeingültigkeit des Ansatzes demonstrieren wir, indem wir ein exemplarisches Agentenmodell in drei weit verbreitete Simulationsumgebungen integrieren: OpenPASS, CARLA und CarMaker.
    Die Evaluation zeigt, dass das Modell auf allen Plattformen ein konsistentes Verhalten aufweist und damit Interoperabilität, Modularität und Standardkonformität der Architektur bestätigt. Unsere Referenzimplementierung reduziert Integrationsaufwände, schafft eine Grundlage für zukünftige Forschung und ist auf GitHub öffentlich verfügbar: \href{https://github.com/ika-rwth-aachen/agent-model-integration}{github.com/ika-rwth-aachen/agent-model-integration}.
    }

    \transkeywords[german]{Simulation, Architektur, Standardisierung, Szenarienbasiertes Testen, Agentenmodell}
    
    \startpage{1}
    \aop

\maketitle

\section{Introduction} 
\label{sec:introduction}

The rapid advancement of automated driving systems necessitates new methodologies to evaluate their safety and performance~\cite{Watzenig2016}.
Since random field tests require extremely large driving distances under real-world conditions to achieve statistical significance, research increasingly focuses on scenario-based testing~\cite{Amersbach2019}.

One key approach to address the large number of safety-critical scenarios is simulation-based testing. Compared to real-world tests, simulation enables cost-effective, deterministic, and automated evaluation. The goal is to reduce complexity while still modeling the environment and the system under test with sufficient fidelity to obtain reliable results. This includes vehicle and sensor models and also naturalistic surrounding traffic, which is commonly represented by parameterizable agent models. Consequently, researchers can assess automated driving systems in complex and realistic interactions with other traffic participants~\cite{hallerbach2018simulation}.

\begin{figure}[htbp]
  \centering
  \includegraphics[width=0.8\columnwidth]{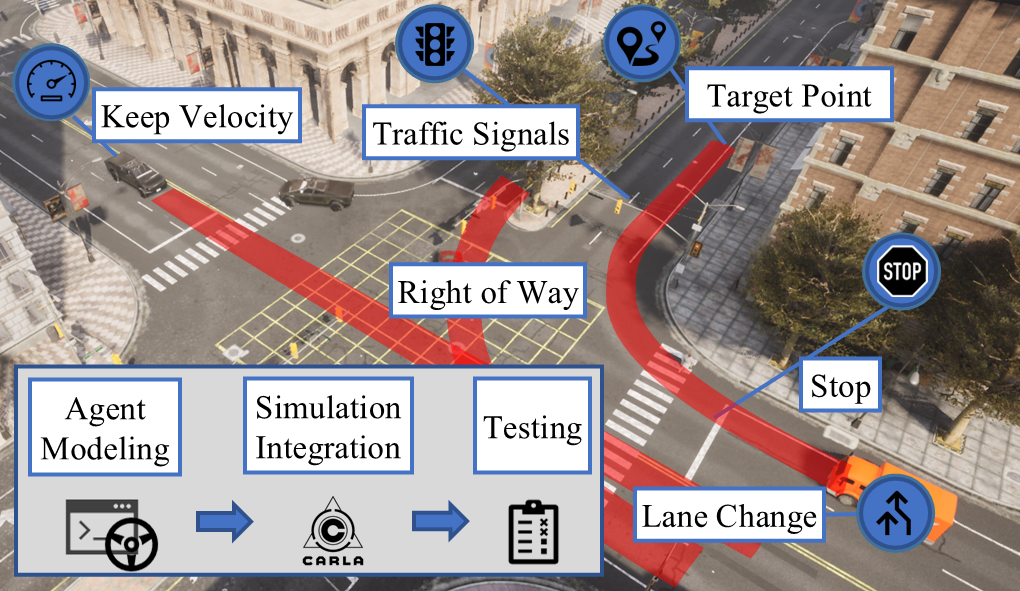}
  \caption{Intersection scenario involving multiple agents within the simulation tool CARLA. The agent model's main capabilities are highlighted to demonstrate a responsive and human-like behavior. The general integration process of a simulation model is illustrated, starting with the development of the agent model, subsequent simulation integration, and subsequent testing.}
  \label{fig:carla-scenario}
\end{figure}

However, the integration of simulation models into different simulation tools remains a technical but also methodological challenge. Existing simulation environments differ in architecture, interfaces, and supported standards. This heterogeneity complicates model reuse, comparison, and verification across platforms. Therefore, harmonized and standardized integration architectures are essential to enable consistent, tool-independent testing workflows. ASAM standards such as the \gls{osi} are promising enablers, yet their application lacks structured implementations usable for further research and development.

This work addresses that gap by providing a reusable, open-source reference implementation that demonstrates how established standards can be combined into a practical, tool-independent integration workflow for closed-loop traffic agent models. Our approach builds on \gls{osi} for structured data exchange and \gls{fmi} for model encapsulation and is realized using the \gls{osmp} framework. The architecture is not limited to a specific simulator or model type but is designed to be reusable across a wide range of simulation-based testing applications.

To demonstrate applicability, we integrate an exemplary, naturalistic, and responsive agent model into three widely used simulation tools, namely OpenPASS, CARLA, and CarMaker. Even if the agent model itself is not the focus, it serves as an illustrative use case to validate interoperability and modularity of the proposed integration approach. 

Our reference implementation is made publicly accessible on GitHub\footnote{Reference implementation is publicly available at \url{https://github.com/ika-rwth-aachen/agent-model-integration}.} and provides a strong foundation for further development and research within the field of simulation-based testing and safety assurance. An overview of the approach is also given in Fig.~\ref{fig:carla-scenario}.

\noindent The main contributions can be summarized as follows:
\begin{itemize}
    \item Generic and standards-compliant simulation integration architecture;
    \item Reusable, open-source reference implementation using \gls{osi} and \gls{fmi};
    \item Integration of a naturalistic, responsive agent model into three simulation tools, OpenPASS, CARLA, and CarMaker;
    \item End-to-end testing framework, providing a comprehensive evaluation of the architecture and the integrated model.
\end{itemize}

\section{Background and Related Work}
\label{sec:sota}

This section provides background on simulation-based testing and traffic agent modeling and reviews related work on standardized simulation interfaces and model-integration approaches relevant to our reference implementation.

\subsection{Simulative Testing} 
\label{sec:sota:simulation-architecture}

Real-world and proving ground testing is essential to verify systems that operate in real-world conditions.
However, applying specific scenarios with various parameter variations is complex and restricted due to boundary limitations~\cite{Amersbach2019,Schütt2022}. In addition, the exact selection of relevant scenarios is a further challenging step~\cite{Safety2019}.Simulations offer another promising way to run many scenarios in a deterministic and cost-effective way. Complexity can be reduced but has to be realistic to obtain meaningful test results. Thus, simulation-based testing is often the first stage in a verification and validation process before selected test cases are performed in a real-world environment~\cite{Steimle2022}.

\begin{figure}[htbp]
    \centering
    \includegraphics[width=0.8\columnwidth]{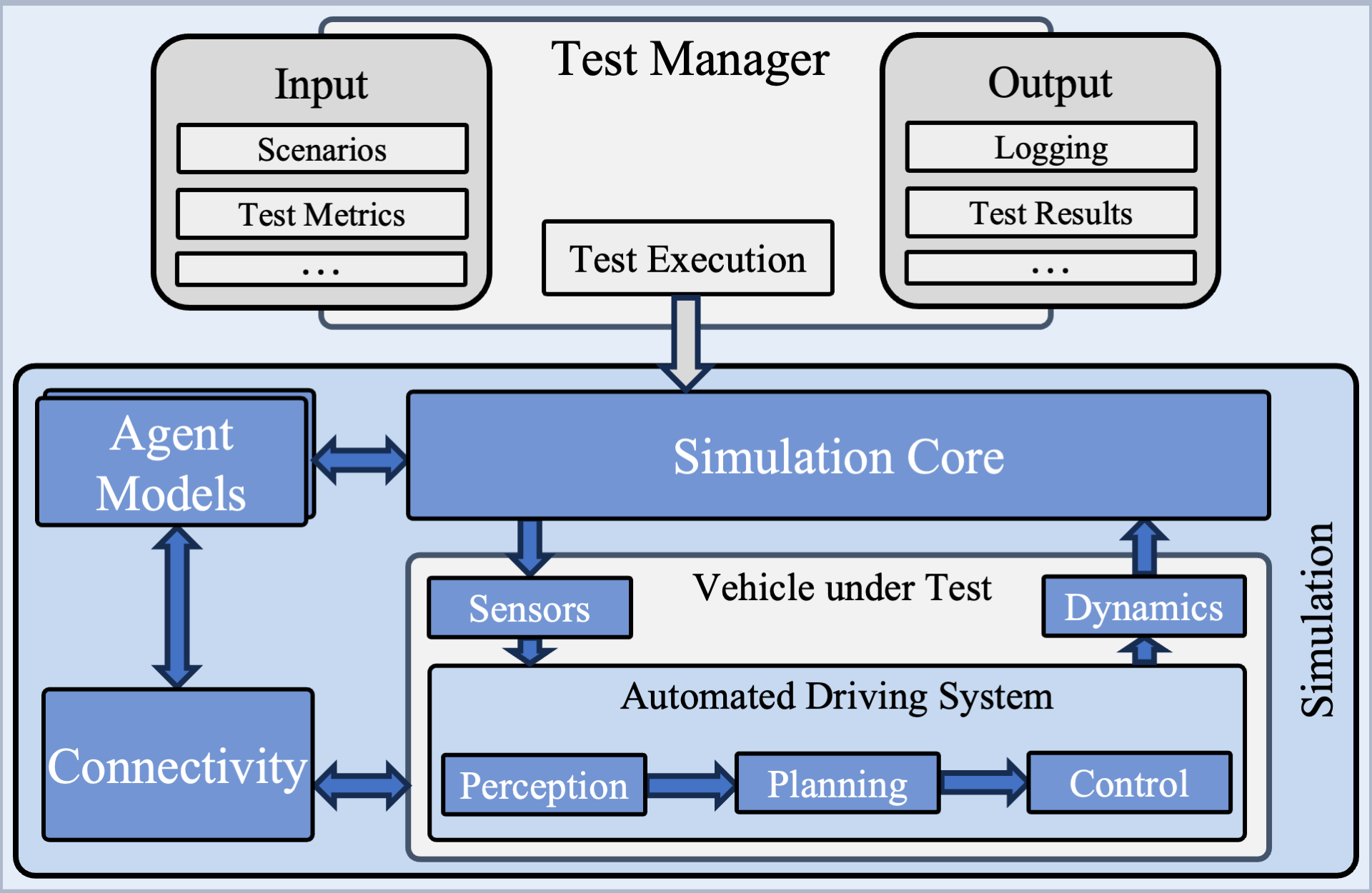}
    \caption{High-level simulation architecture for scenario-based testing of automated driving systems. Every simulation is configured using a scenario and a test description. Logs and test results are stored as simulation output. Multiple components are involved in a simulation, where the vehicle under test with an automated driving function receives the highest level of detail. The surrounding traffic is modeled with less fidelity but as naturalistic as necessary using multiple closed-loop agent models. (Figure adapted from~\cite{SunriseDeliverable2024}.)}
    \label{fig:simulation-overview}
\end{figure}

While the vehicle under test is typically modeled in high detail using a complex automated driving function, other road users are often represented by simplified agent models. Fig.~\ref{fig:simulation-overview} illustrates the interaction between the simulation core, the vehicle under test, and additional simulation components such as agent models. Furthermore, test execution and evaluation are handled by a dedicated test manager.

We focus on the integration and interaction between simulation models and the simulation core. The approach is demonstrated for a traffic agent model, but it can be extended to sensor or vehicle models as well. Therefore, the following sections provide background on both simulation integration techniques and relevant aspects of agent-based traffic modeling.

\subsubsection{Simulation Tools} 
\label{sec:sota:simulation-tools}

The simulation tools in the context of automated driving are differentiated primarily by their level of detail and focus. Since tool suitability depends on the intended application, we provide a brief overview from an integration perspective rather than an exhaustive tool comparison. Nevertheless, simulation models such as agent, sensor, or vehicle models should be integratable across different simulation tools.

{IPG CarMaker}~\cite{Ipg2026} originates from the vehicle dynamics perspective and provides a very accurate vehicle model. Scenario-based testing, as well as custom model integrations, are also key functionalities~\cite{Ueberbacher2017}.

The open-source simulation tool {OpenPASS}~\cite{Dobberstein2017} focuses on scenario-based testing in the context of safety assurance. The main objectives are a realistic traffic simulation, the stochastic variation of scenarios, and simple and modular interfaces for the integration of custom models~\cite{Dobberstein2017}.

In order to achieve a realistic environment representation, {Virtual Test Drive}~\cite{Hexagon2026} and the open-source simulator {CARLA}~\cite{Carla2017} represent further possibilities in the context of automated driving simulations~\cite{Carla2017}. These tools primarily enable detailed sensor modeling and visualizations in combination with scenario-based testing strategies~\cite{Geller2024}. {CARLA}, in particular, offers large community support and enables a wide range of third-party modules and interfaces.

In addition, microscopic traffic simulators, such as the open-source {SUMO}~\cite{Lopez2018} simulator, are widely used to generate large-scale traffic flows and to parameterize traffic demands.

\subsubsection{Simulation Interfaces} 
\label{sec:sota:simulation-interfaces}
Using standardized interfaces simplifies the exchangeability of models across simulation tools. {ASAM~e.V.} is a major standardization organization in the simulation domain that aims to harmonize development and testing processes in the automotive context~\cite{Asam2026}.

There are {ASAM} standards for the specification of the scenario and the map environment, such as {OpenSCENARIO} and {OpenDRIVE}. Another standard, the \gls{osi}~\cite{Osi2026}, focuses on exchanging data between different simulation components aligned with the ISO 23150 standard. \gls{osi} ultimately describes both the ground truth information, such as object information or lane layouts, and sensor data containing the output of a modeled sensor. In addition, there are interfaces to interact with traffic agent models. These include traffic commands to pass destination points or a desired speed. In return, the traffic agents update their pose as \gls{osi} message. Finally, all data is structured in an object-oriented architecture using Google's Protobuf message format, enabling an efficient and standardized way of serializing data~\cite{Asam2026}.

A different standard for simulation is the \gls{fmi}~\cite{Fmi2026}, which aims to integrate different software components. Thus, individual simulation components can be exchanged as a \gls{fmu}. \gls{fmu}s transfer data as scalar input and output variables defined in a specification file. Many existing tools such as {OpenPASS}, {SUMO} or {CarMaker} already offer \gls{fmi} support. Additionally, third-party open-source solutions such as OSTAR~\cite{Ostar2024} enable \gls{fmi} for tools like {CARLA}.

\subsubsection{Simulation Integration Approaches}
\label{sec:sota:integration-approaches}

A variety of approaches exist to integrate sensor or agent models into simulation environments, differing in abstraction level, generality, and standardization.

One standardized approach is the \gls{osmp} project~\cite{Osmp2026}, which combines the object-oriented \gls{osi} with the scalar-based \gls{fmi}. \gls{osmp} enables efficient runtime performance by transferring only pointers in \gls{osi} messages using shared memory, while retaining modularity through \gls{fmi} packaging. This allows for the development of \gls{fmu}-based models that access \gls{osi} data structures directly during runtime. Several implementations of \gls{osmp}-compatible \glspl{fmu} have been published, including sensor models~\cite{Rosenberger2020,Linnhoff2020} and traffic agent models~\cite{Lemmer2021,Bartolozzi2022}. A list of open-source \gls{osmp}-capable models is maintained in the ENVITED Model \& Simulation Library~\cite{Envited2026}.

Alternative integration frameworks include MECSYCO (Multi-agent Environment for Complex SYstem CO-simulation)~\cite{Mecsyco2016} and the High Level Architecture (HLA)~\cite{Ieee1516}. MECSYCO builds on the \gls{devs} formalism to couple heterogeneous simulation models, including both discrete-event and continuous-time systems. While MECSYCO offers flexibility and supports distributed simulation, it lacks native support for automotive domain-specific standards like \gls{osi}, which are essential in the context of automated driving~\cite{Mecsyco2016}. HLA, standardized as IEEE 1516~\cite{Ieee1516}, was originally developed for coordinating distributed simulation systems in defense. It provides formal interfaces and a runtime infrastructure for synchronization and data exchange. Despite its general applicability, HLA has been applied to automotive simulation in some research contexts~\cite{Reiher2024,Neema2024}, but it remains complex to implement and has seen limited application in automated driving due to the absence of domain-specific interfaces.

Other solutions for the integration of simulation models rely on middleware technologies or direct communication interfaces. Examples include \gls{ros}-based systems, TCP/IP messaging, or custom shared memory implementations. These interfaces are often used in research settings for specific tool integrations. Although they allow for flexible setup and fast prototyping, such approaches are usually tailored to individual use cases, limiting their portability and reusability.

Many simulation tools also offer tool-specific interfaces for model integration. For example, CARLA supports integration and control via C++ and Python APIs~\cite{Carla2017}; OpenPASS relies on custom-built models in C++~\cite{Dobberstein2017} and CarMaker provides a C/C++ interface for integrating custom traffic or vehicle models. These integrations provide deep access to the simulator internals, but are tightly coupled to the respective simulation environment, limiting reuse across platforms.

\newpage
In summary, these various approaches illustrate the diversity of simulation integration strategies in current use. They range from flexible but tool-specific solutions to formalized, system-level co-simulation frameworks. The extent to which they enable reusable, modular, and standards-compliant model integration varies significantly depending on their architecture and intended application context.

\subsection{Driver Behavior Modeling} 
\label{sec:sota:driver-behavior-modeling}

Driver behavior models have been the subject of extensive studies in the last decades to better mimic driver behavior in simulations. However, most research in this area usually focuses on one of the three layers of the driving task: navigation, guidance, or stabilization~\cite{mcruer_newresults}. Navigation is the cognitive process of planning and monitoring a vehicle's route, while guidance concerns the more immediate path planning. The stabilization task enables guidance by feeding inputs into the vehicle dynamics model in the form of acceleration and steering commands. In this contribution, we are mainly concerned with the guidance and stabilization layers. The stabilization layer contains models that are either concerned with the lateral or longitudinal domain. Other behavior models capture the human decision-making process and fall under the guidance task. The output of a guidance model is then usually used as an input for stabilization.

A commonly used model to describe movement in the longitudinal domain is the intelligent driver model~(IDM) presented by Treiber, Hennecke, and Helbing~\cite{treiber2000congested}. This model describes changes in the speed of a vehicle based on several factors, such as the desired velocity, minimum acceptable distance to a vehicle in front, the desired \gls{thw}, and values for maximum acceleration and braking. These variables are used in a series of differential equations to calculate the momentary acceleration of the vehicle.

For lateral control, Salvucci and Grey proposed a control scheme based on two points on the desired path~\cite{salvucci2004two}. The goal of the scheme is to minimize the angle between the heading of the vehicle and the direction to the points on the track by using a PID controller. The use of two points allows both the stabilization of the vehicle on the path as well as some ability to react to upcoming changes in the path, allowing for more realistic human behavior.

To accurately describe driver behavior, it is necessary to use lane change models that consider the decision-making process. The lane change decision can be divided into three levels: strategic, tactical, and operational~\cite{kesting2007general}. At the strategic level, decisions regarding lane changes are made to complete the desired route. The tactical level is used to prepare for the actual lane change, while the operational level considers whether the lane change can be performed safely.

Kesting, Treiber, and Helbing developed a model specifically for the operational layer of the lane change decision-making process \cite{kesting2007general}. The decision of whether a lane change can be made safely depends on several criteria, including a safety gap, low relative velocities, as well as the impact of the lane change on other drivers. In addition, other criteria can be used to incentivize legally required behavior, such as passing on the left and keeping to the rightmost lane, as mandated in many jurisdictions with right-hand traffic.

One can now aggregate models that cover all different parts of the driving task and describe driving behavior as a whole. The result is a traffic agent that can be used to traverse diverse traffic environments. Examples of such traffic agents are the Stochastic Cognitive Model~\cite{fries2021validation} and DReaM~\cite{Siebke2022}, as well as the representative agent model used in this work.

To use a driver behavior model within a simulation tool, the model requires access to the current simulation state, including other agents and static elements such as the road network. In return, the model provides control outputs for its own agent, which are then used to update the vehicle state. This bidirectional exchange is required for any combination of driver behavior model and simulation tool.

\section{Methodology \& Implementation}
\label{sec:implementation}

This section outlines the methodological steps and implementation details used to realize and validate the proposed reference implementation. The methodology consists of three stages:
\begin{itemize}
 \item The implementation of a standardized, modular integration architecture based on \gls{osi} and \gls{fmi};  
 \item The integration of a representative closed-loop simulation agent model;  
 \item The evaluation of both the architecture itself and the resulting model behavior.
\end{itemize}

The goal is to enable standardized, reusable integration of closed-loop simulation models across heterogeneous simulation tools. The process involves the design of a modular architecture, its implementation using the \gls{osmp} framework, and its application using a representative agent model.

While the agent model itself is not the focus of this work, it serves as an illustrative and parameterizable example to validate the integration concept in practice. It enables the demonstration of runtime consistency, portability, and seamless reusability across tools such as OpenPASS, CARLA, and CarMaker. The overall approach, from agent modeling to simulation integration and subsequent scenario-based testing, is summarized in Fig.~\ref{fig:carla-scenario} and detailed in the following subsections.

The first step involves the development of a responsive and parametrizable agent model that serves as a use case for demonstrating the integration process (cf. Sec.~\ref{sec:implementation:agent-model}). This model simulates human-like driving behavior in a closed-loop setup. The next step is the closed-loop integration of the model into an overall simulation architecture. Standardized interfaces and modular architectures are essential, both of which can be realized using the \gls{osmp} framework (cf. Sec.~\ref{sec:implementation:osmp}).
Finally, the integration enables the automated test execution of various scenarios to validate the agent model's capabilities according to predefined metrics (cf. Sec.~\ref{sec:implementation:simulative-testing}).

To assess whether the proposed integration architecture fulfills its intended purpose, we define four evaluation criteria derived from core requirements in simulation-based testing:

\begin{itemize}
  \item \textbf{Standards Compliance:} The integration approach should rely on widely accepted simulation standards and avoid tool-specific or proprietary interfaces.
  \item \textbf{Portability and Reusability:} Simulation models should be technically transferable across different tools without requiring modifications or reimplementation.
  \item \textbf{Integration Effort:} The effort required to integrate a model into a new simulation environment should be minimal, ideally without additional development or manual adaptation.
  \item \textbf{Result Consistency:} Under identical test conditions and inputs, the same model should exhibit functionally equivalent behavior across simulation environments, indicating interoperability.
  \item \textbf{Scalability:} The integration approach should scale efficiently with the number of coupled model instances in terms of runtime and memory consumption.
\end{itemize}

The selected criteria focus on the core contribution of this work, namely the tool-independent integration of closed-loop simulation models. Standards compliance and portability capture the ability to exchange models across simulators, integration effort reflects the practical feasibility of adopting the approach, and result consistency assesses whether the coupled model yields plausible behavior across different simulation environments under identical inputs.

Additional criteria become relevant advanced simulation use cases, such as real-time performance, determinism, logging, and traceability. Here, we limit the evaluation scope to the integration concept and its tool-independent realization. Many of the additional criteria depend strongly on the simulator core, its configuration, and the runtime and hardware setup, and are therefore treated as complementary aspects rather than primary evaluation targets.

Therefore, the listed criteria above are used in Sec.~\ref{sec:evaluation:architecture} to evaluate the reference implementation. In addition, Sec.~\ref{sec:evaluation:model} illustrates the runtime behavior of the integrated agent model, and Sec.~\ref{sec:evaluation:scalability} assesses scalability in terms of runtime and memory consumption.

\subsection{Agent Model} 
\label{sec:implementation:agent-model}

\begin{figure}[htbp]
    \centering
    \includegraphics[width=0.8\columnwidth]{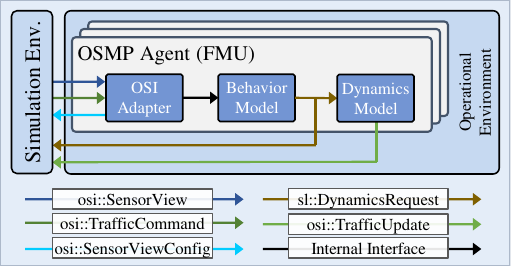}
    \caption{Agent model packed as \gls{fmu} integrated into an \gls{osi}-based simulation architecture. The model is separated into three parts which are a sensing layer~(\gls{osi} adapter) that interprets the ground truth information of an \texttt{osi3::SensorView}, a behavior model that outputs a custom \texttt{sl::DynamicsRequest} message, and a dynamics model which transforms desired values into a \texttt{osi3::TrafficUpdate}. Optionally, the output of the behavior model can be returned directly to the simulation.}
    
    \caption*{\footnotesize\textit{Note:} The custom \texttt{sl::DynamicsRequest} has been officially standardized in \gls{osi} and is publicly available as \texttt{osi3::MotionRequest}. This standardization of \gls{osi} was actively driven by our work and implementation. Migration is planned once the new \gls{osi} version is fully supported across all relevant simulation tools.}
    \label{fig:architecture}
\end{figure}

As a use case for the proposed integration approach, we present a responsive agent model that imitates human-like driving behavior and can be coupled to a closed-loop simulation architecture using the \gls{osi} standard. The model's components are visualized within the white box shown in Fig.~\ref{fig:architecture} and are detailed in this section.

For the translation and interpretation of \gls{osi} data, an \gls{osi} adapter has been developed as part of the agent model, which processes the standardized input messages \texttt{osi3::SensorView} and \texttt{osi3::TrafficCommand}. The \gls{osi} adapter not only performs a simple transformation but also interprets the \gls{osi} data, thus representing a sensing layer for the agent model.

As an example, the target position as well as the desired speed values are extracted directly from the \texttt{osi3::TrafficCommand}. Combined with the lane layout from the \texttt{osi3::SensorView}, a Dijkstra-based~\cite{Dijkstra1959} route planning algorithm calculates the lane path needed to reach the desired destination. On the sensing layer, the \gls{osi} wrapper interprets information about traffic signals or surrounding traffic relevant to the current position and converts them into a local virtual horizon representation. 

In addition, the lane layout is analyzed regarding necessary lane changes. In order to find a safe lane change position, potential conflict zones such as intersections or solid lines are considered to optimize the lane change behavior.

This preprocessing is used to populate the internal interface of our agent model~\cite{SimDriver2026}. The remainder of the model is divided into a driver behavior and a subsequent dynamics model. The former computes values for longitudinal and lateral movement represented by a desired acceleration $a_{des}$ and a desired curvature $\kappa_{des}$, respectively. These quantities are encapsulated in a custom \gls{osi} message \texttt{sl::DynamicsRequest}, which has since been standardized as official \gls{osi} message within the \texttt{osi3::MotionRequest}. The dynamics model then transforms the desired quantities into the vehicle's actual movement, which is returned to the simulation environment within the \texttt{osi3::TrafficUpdate} message. Both parts are further described in the following sections.

\subsubsection{Behavior Model} 
\label{sec:implementation:agent-model:behavior-model}
The driver behavior model is based on the theoretical foundations described in \cite{Klimke2020}. Table~\ref{table_capabilities} summarizes the basic capabilities required to realize typical maneuvers, such as reaching a target position within a scenario. For example, reaching a point behind a signalized intersection in a different lane requires coordinated decisions on traffic light handling, road-user interactions, and lane changes.

These decisions are carried out by associating them with distinct guidance values, which can be a desired velocity or a distance along the virtual horizon to the next stop point. These guidance values are then controlled by the two desired stabilization variables $a_{des}$ and $\kappa_{des}$. This process follows the driver modeling approach of Donges~\cite{Donges1978}.

\begin{table}[htb]
\caption{Selected behavior model capabilities with corresponding guidance values and short description.}
\label{table_capabilities}
    \begin{center}
        \begin{tabularx}{\linewidth}{cYY}
            \toprule
            \textbf{Capability} & \textbf{Guidance Values} & \textbf{Description} \\
            \midrule
                    
            Free Driving & Desired velocity, speed limit, curve speed (dependent on lateral acceleration) & 
            \vspace{0.33cm}
            \begin{tabitemize}
                \item Predictive velocity adaptation
                \item IDM-based reaction
                \item Smooth acceleration 
            \end{tabitemize}\\
            \midrule            
            Following & Desired time headway~(THW) & 
            \vspace{0.33cm}
            \begin{tabitemize}
                \item IDM-based adaptation
                \item Smoothly reach desired THW
                \item Stopping is based on the same reaction but parameterized differently
            \end{tabitemize} \\
            \midrule            
            Lane Keeping & Reference points on virtual horizon & 
            \vspace{0.33cm}
            \begin{tabitemize}
                \item Behaves similarly to a PID controller
                \item Based on Salvucci and Grey steering model~\cite{salvucci2004two}
            \end{tabitemize} \\
            \midrule
            Lane Change & Desired lateral offset &
            \vspace{0.33cm}
            \begin{tabitemize}
                \item Dijkstra route planning~\cite{Dijkstra1959}
                \item Decision based on traffic and junction layout
                \item Dynamic lateral offset fading to new lane centerline by interpolation
            \end{tabitemize} \\
            \bottomrule
        \end{tabularx}
    \end{center}
\end{table}
 
The longitudinal behavior modeling is based on a slightly modified IDM (cf. Sec.~\ref{sec:sota:driver-behavior-modeling}). In general, the adjusted IDM equation for the acceleration may be formulated as follows:
\begin{equation}
	a_{des} = a_{max} (1 - \sum_{i \in R} r_i),\, R = \{\textrm{free, stop, follow}\}.
	\label{eq:idm}
\end{equation}

The maximum applicable acceleration is denoted by $a_{max}$. The actual desired acceleration $a_{des}$ is influenced by three reaction terms for free driving, stopping, and following, respectively. These reactions are always positive, which results in desired accelerations of at most $a_{max}$ and possibly large deceleration values, e.g., to stop immediately for a cut-in of another target. Note that the underlying dynamics model might bound the actual desired deceleration value.

With $\Delta v = v_T - v$, the term $r_{\text{free}}$ (cf.~(\ref{eq:idm_free})) is formulated diverging from the original IDM. Here, $v_T$ denotes the target velocity provided by the guidance layer, and $v$ is the current vehicle speed. In contrast to the original ratio-based term $(v/v_T)^\delta$, which primarily models the reduction of acceleration when approaching $v_T$ from below, our deviation-based formulation explicitly captures both undershoot and overshoot around the desired speed with symmetric sensitivity. This enables applying the same reaction magnitude for equal absolute deviations from $v_T$ while still distinguishing acceleration from deceleration depending on the sign of $\Delta v$:

\begin{equation}
	r_{\text{free}}(\Delta v) =
	\begin{cases}
	\left(1-\frac{|\Delta v|}{v_T}\right)^\delta, & \Delta v \ge 0\\
	2-\left(1-\frac{|\Delta v|}{v_T}\right)^\delta, & \Delta v < 0
	\end{cases}
	\label{eq:idm_free}
\end{equation}

The desired speed $v_T$ is influenced by the factors listed in Table~\ref{table_capabilities}. 
According to \cite{treiber2000congested}, the parameterizable exponent $\delta$ is usually set to 4. 
To enable predictive speed adaptation, the free-driving reaction is evaluated not only at the vehicle's current position but also along a virtual look-ahead horizon. 
The horizon length is defined as $ds_{\max}=vT_{\max}$, i.e., it scales linearly with the current speed and is therefore bounded in time rather than space. 
In implementation, this horizon is represented by a discrete set of preview points $ds_i \in [0, ds_{\max}]$, at which upcoming speed constraints are evaluated. 
These values are combined into a predictive target speed $v_{T,\mathrm{pred}}$, and we then select the more conservative of the local and predictive free-driving reactions. This yields anticipatory deceleration in response to a speed-limit sign or an upcoming curve, similar to most human drivers.

A common equation describes the reactions for stopping and following:
\begin{equation}
	r_{follow/stop} = \left(\frac{ds^*}{ds}\right)^2,\, ds^* = s_0 + v T + \frac{v(v-v_{\text{pre}})}{2 \sqrt{a b}}. 
	\label{eq:idm_interaction}
\end{equation}
In (\ref{eq:idm_interaction}), $s_0$ denotes the net distance to a stopping point or a leading vehicle. $T$ describes a desired \gls{thw} and is treated as a constant in both modes. In following mode, $T$ directly represents the constant desired \gls{thw} to the leading vehicle. In stopping mode, the same constant $T$ can be used as a tuning parameter to shape a degressive approach behavior. The preceding vehicle's velocity is given by $v_{\text{pre}}$, and finally, $a$ and $b$ denote tunable parameters for acceleration and deceleration for the maneuvers following and stopping. In implementation, we compute an effective headway $T_{\mathrm{eff}} = T - \frac{s_0}{v_{\mathrm{pre}}}$. This compensates the standstill offset $s_0$ so that the measured \gls{thw} matches the specified \gls{thw}. The corrected headway is then used in the following and stopping reaction term.

The separate calculation of longitudinal reactions enables high modularity and achieves distinct, parameterizable behavior within the model. In addition, this allows us to calibrate and test each capability according to human-like behavior, e.g., captured in measurement data.

The lateral movement is inspired by the two-point visual control model of steering introduced in \cite{salvucci2004two} (cf. Sec.~\ref{sec:sota:driver-behavior-modeling}). A near and far reference point along the virtual horizon are selected, and the angular deviation to both is minimized by calculating a desired curvature $\kappa_{des}$. The longitudinal distance of the selected reference points is speed-dependent and rises with increasing speed. In lateral direction, the points are, by default, located on the lane's centerline. However, they can be manipulated by a parameter, e.g., to model a driver that constantly drives with an offset to the left or to model a lane change continuously without an abrupt relocation of reference points to the neighboring lane. To compensate for oscillations, the time derivative of the angular change is also considered.

For further implementation details, we refer to the open-source repository of our agent model~\footnote{\url{https://github.com/ika-rwth-aachen/SimDriver}}.

\subsubsection{Dynamics Model} 
\label{sec:implementation:agent-model:dynamics-model}

Once the desired control variables of the behavior model are calculated, two options emerge for using them in a simulation environment. One option involves the direct transmission of the \texttt{sl::DynamicsRequest} message to the simulation environment so that desired acceleration and curvature are used within an integrated dynamics module. Alternatively, the dynamics module of the agent model itself can be used.

The dynamics model's goal is to translate the desired outputs of the behavior model into an updated vehicle pose that can be set within the simulation environment. The model includes two PID controllers that calculate a pedal value from $a_{des}$ and a steering angle from $\kappa_{des}$, respectively. In addition, a single-track model, as part of the dynamics module, computes an updated vehicle state based on the pedal and steering values. The resulting pose update is then embedded in an \texttt{osi3::TrafficUpdate} message, constituting the model's overall output.

\subsection{OSI-based Simulation Integration} 
\label{sec:implementation:osmp}
When integrating simulation models into a simulation environment, challenges often arise due to differing interface paradigms, simulator-specific APIs, or inconsistent support for standardized data formats. To address these issues, we propose a modular integration architecture based on the ASAM \gls{osi} and \gls{fmi} standards. The implementation is realized using the \gls{osmp} framework and was primarily developed in the SET Level project~\cite{Setlevel2026}. An overview of the architecture is provided in Fig.~\ref{fig:architecture}.

Within this setup, the agent model, including all functional components, is encapsulated as an \gls{osmp}-compatible \gls{fmu}. This enables a standardized interface between the simulation model and the simulation core. One central challenge is the efficient coupling of \gls{osi}'s object-oriented Protobuf structures with \gls{fmi}'s scalar-based co-simulation mechanism. \gls{osmp} resolves this by using shared memory: Protobuf message pointers are exposed through \gls{fmi} integer variables, avoiding costly data serialization or transformation. In each simulation time step, \texttt{osi3::SensorView} and \texttt{osi3::TrafficUpdate} are exchanged synchronously. Event-driven control commands (e.g., speed targets or waypoints), which are only sent when needed or updated, use \texttt{osi3::TrafficCommand}.

Robustness against edge cases is achieved by leveraging the \gls{fmi} error handling mechanisms on the simulator host side. In addition, \gls{osmp} provides logging during \gls{fmu} instantiation and initialization and all incoming \gls{osi} messages are validated within the agent model's \gls{osi} adapter layer before they are processed in the model itself.

Each road user in a scenario is represented by an independent \gls{fmu} instance. Runtime parameters such as desired and initial speed or debugging flags are provided via \gls{fmi} variables and can be set consistently across different simulation tools. This design supports scalable multi-agent setups and consistent execution across platforms.

A compatible simulation environment needs to be able to (i) instantiate \gls{fmi} components and (ii) generate valid \gls{osi} ground truth data. We successfully integrated the \gls{osmp}-packaged agent model into three simulation tools with varying architectures: OpenPASS, CARLA, and CarMaker.

\subsection{Simulative Testing} 
\label{sec:implementation:simulative-testing}

Once a simulation model is embedded into the proposed architecture, scenario-based testing can be performed consistently across different simulation environments. The standardized \gls{osmp} integration ensures uniform interface semantics, allowing a single set of test scenarios and evaluation functions to be reused without modification.

\begin{figure}[htbp]
\centering

\begin{subfigure}[b]{0.3\columnwidth}
\includegraphics[width=\columnwidth, height=2.5cm]{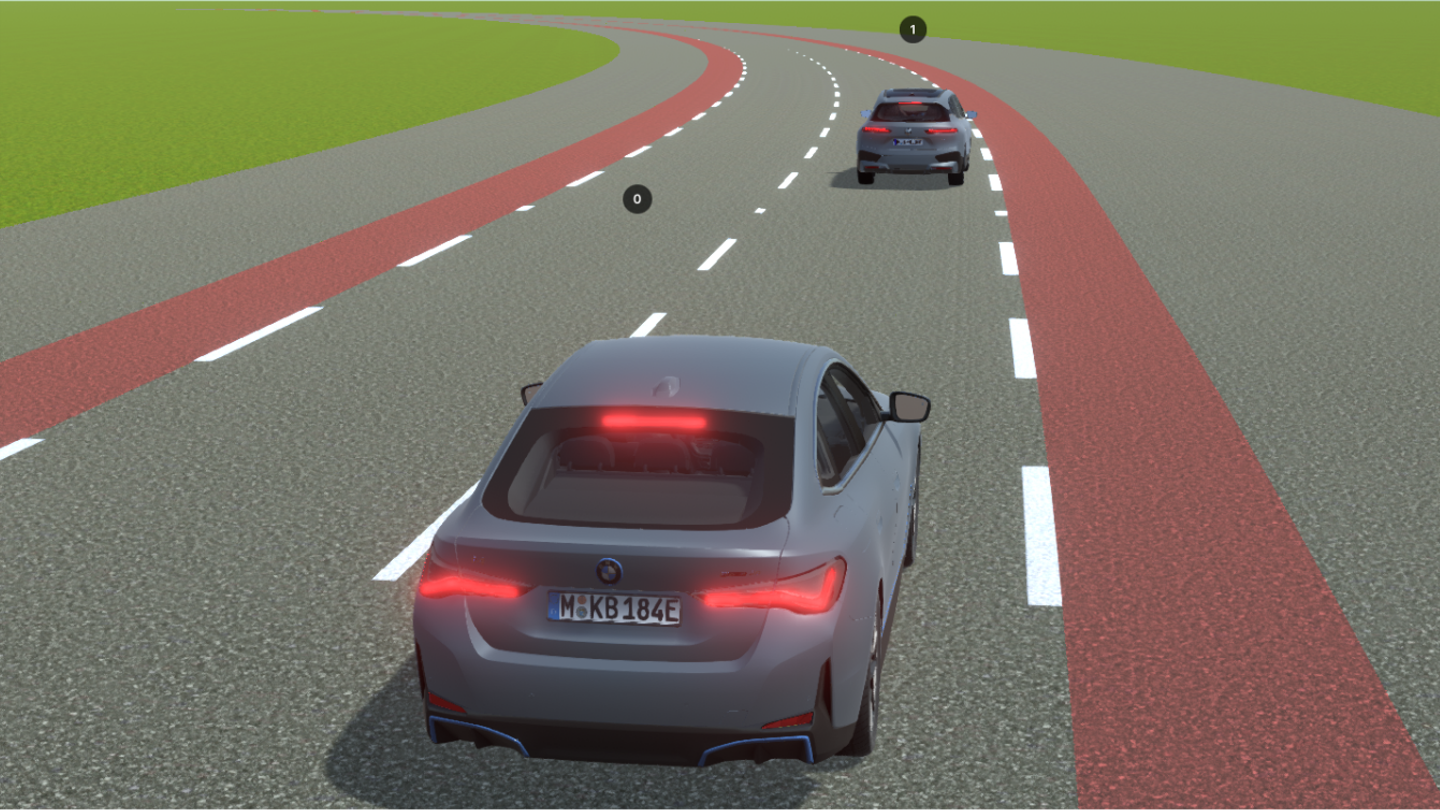}
\caption{OpenPASS}
\end{subfigure}
~
\begin{subfigure}[b]{0.3\columnwidth}
\includegraphics[width=\columnwidth, height=2.5cm]{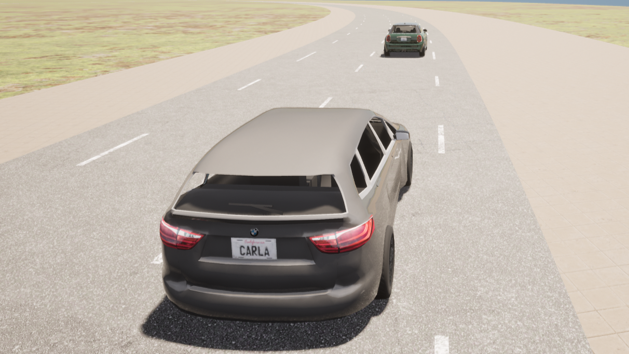}
\caption{CARLA}
\end{subfigure}
~
\begin{subfigure}[b]{0.3\columnwidth}
\includegraphics[width=\columnwidth, height=2.5cm]{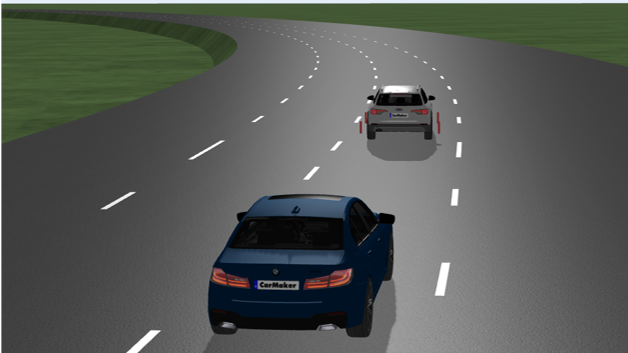}
\caption{CarMaker}
\end{subfigure}

\caption{Exemplary snapshots of a following scenario executed in three simulation tools using the proposed reference implementation for model integration.}
\label{fig:tool-comparison}
\end{figure}

The simulation architecture itself is built around widely adopted standards. Scenario definitions are based on OpenSCENARIO and OpenDRIVE, ensuring tool compatibility with existing simulation platforms. This also applies to the test execution layer, which follows a modular structure and includes a configurable test manager to evaluate predefined metrics.

To illustrate cross-tool interoperability, we execute the same vehicle-following scenario in all three simulation environments. Fig.~\ref{fig:tool-comparison} shows exemplary snapshots from OpenPASS, CARLA, and CarMaker. Beyond behavioral plausibility, these examples highlight the ability to perform standardized and repeatable closed-loop tests across different simulators with minimal integration effort, since the same model and test scenario are used in all three simulation environments.

To structure the evaluation of the integrated agent model itself, we first define a set of model capabilities. We distinguish between \emph{basic} capabilities, which are simulator- and maneuver-agnostic building blocks required in all scenarios, and \emph{specific} capabilities, which are only meaningful in selected scenarios. An overview of all capabilities is given in Table~\ref{tab:capabilities}.

\newcommand{\tabitem}{\textbullet\hspace{0.5em}}

\begin{table}[ht]
\centering
\caption{Categorization of agent model's main capabilities: Basic capabilities represent generic building blocks evaluated across all scenarios, while specific capabilities are only relevant in specific scenarios.}
\label{tab:capabilities}
\begin{tabular}{@{\hspace{0.5cm}}p{0.25\linewidth}@{\hspace{0.5cm}}p{0.25\linewidth}@{}}
    \toprule
    \textbf{Basic Capabilities} & \textbf{Specific Capabilities} \\
    \midrule
    \tabitem Target point        & \tabitem Lane change \\
    \tabitem Desired velocity    & \tabitem Curve handling \\
    \tabitem Lateral offset      & \tabitem Following \\
    \tabitem Yaw offset          & \tabitem Right of way \\
    \tabitem Acceleration limits & \tabitem Traffic signs \\
    \bottomrule
\end{tabular}
\end{table}

Based on these capabilities, we assembled a test catalog that covers the considered behaviors with corresponding pass/fail criteria. The catalog comprises 13 scenarios with a runtime of \SIrange{30}{60}{\second} per run and is specified using OpenSCENARIO and OpenDRIVE formats. This enables executing the same catalog in all three considered simulation tools. Exemplary evaluation metrics include a minimum allowed \gls{thw} to a preceding vehicle, bounded longitudinal acceleration when reaching a speed, and limited lateral acceleration during lane changes. Additional criteria cover lateral deviation, yaw offset, collision counts, and route progress with goal completion.

\begin{figure}[htbp]
\centering

\begin{subfigure}[b]{0.3\columnwidth}
\includegraphics[width=\columnwidth, height=2.5cm]{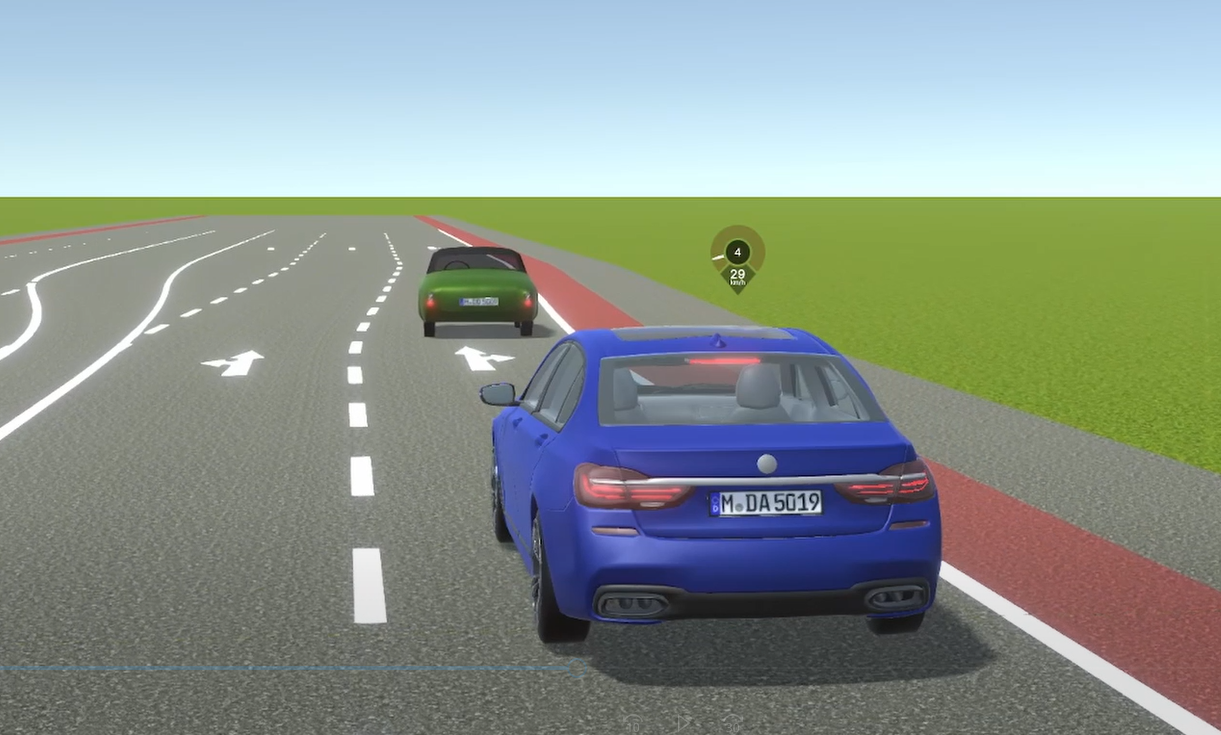}
\caption{Following}
\end{subfigure}
~ 
\begin{subfigure}[b]{0.3\columnwidth}
\includegraphics[width=\columnwidth, height=2.5cm]{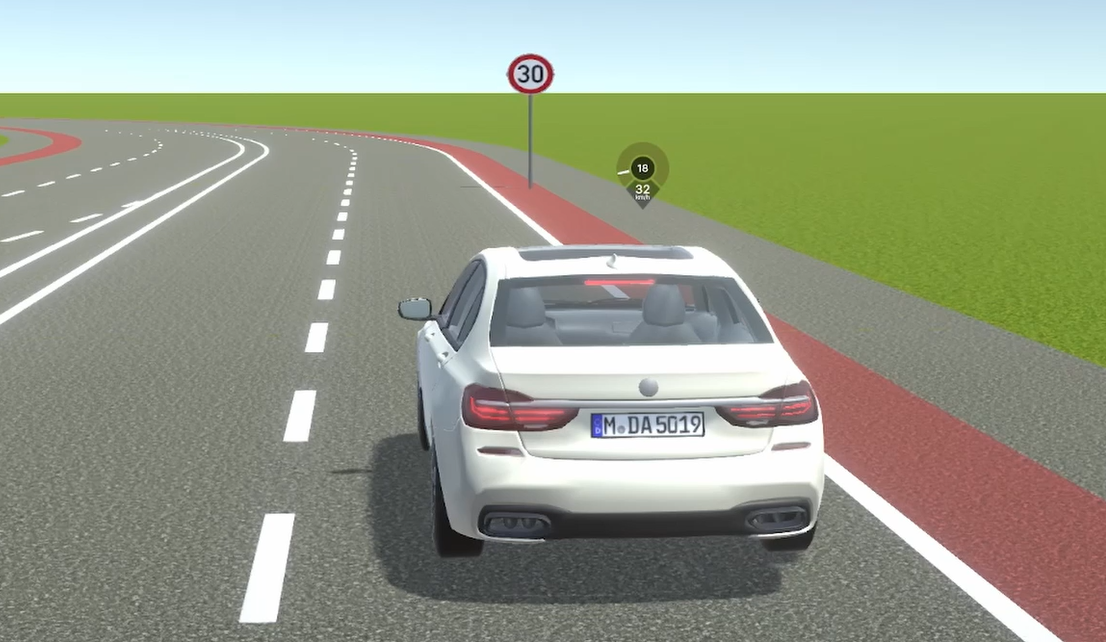}
\caption{Speed Limit}
\end{subfigure}
~ 
\begin{subfigure}[b]{0.3\columnwidth}
\includegraphics[width=\columnwidth, height=2.5cm]{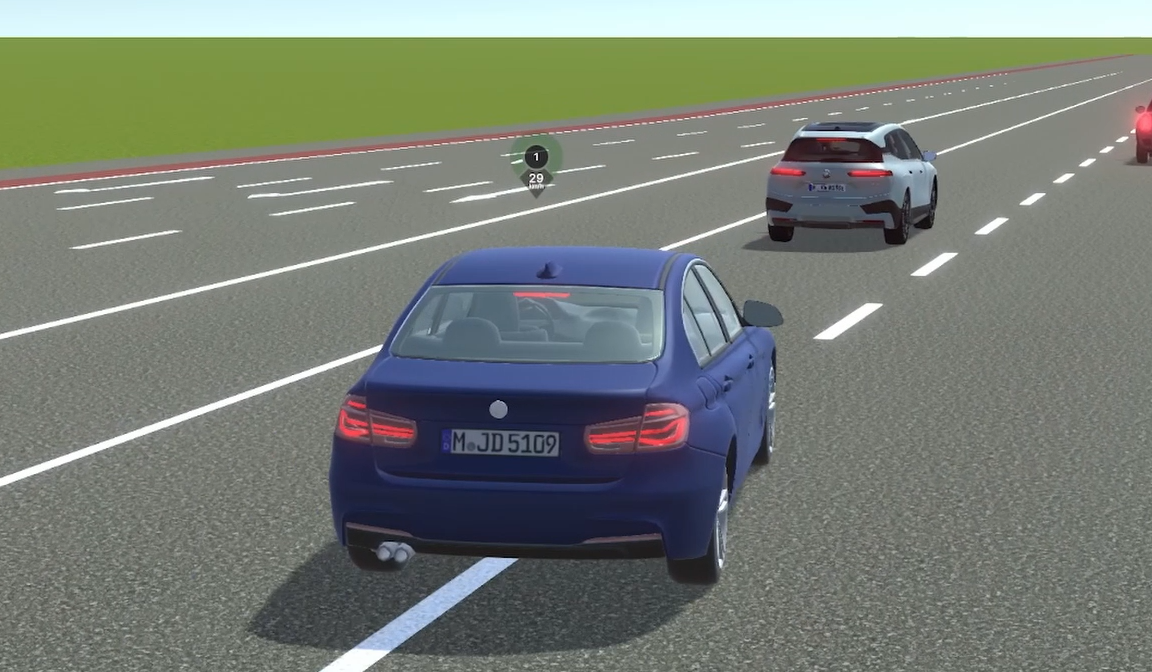}
\caption{Lane Change}
\end{subfigure}

\caption{Representative minimal scenarios from the test catalog, simulated in OpenPASS.}
\label{fig:basic-maneuvers-openpass}
\end{figure}

From this catalog, we select three representative minimal scenarios to assess key driving capabilities in isolation, namely car-following, compliance with speed limits, and tactical lane changes. The scenarios are visualized in Fig.~\ref{fig:basic-maneuvers-openpass} using the OpenPASS visualizer and serve as a basis for the evaluation in Sec.~\ref{sec:evaluation:model}.

Using an end-to-end testing framework and OpenPASS, we implemented an automated testing pipeline to run the well-defined scenarios from the test catalog as part of continuous integration~(CI) jobs. This setup facilitates reproducible verification of the model's core capabilities and supports continuous compatibility checks against the current versions of the supported simulation tools.

\section{Evaluation}
\label{sec:evaluation}

To evaluate the proposed architecture, we consider both the reference implementation itself and its application using a representative agent model. Sec.~\ref{sec:evaluation:architecture} assesses the architecture based on the criteria introduced in Sec.~\ref{sec:implementation}, independent of a specific simulation model. Sec.~\ref{sec:evaluation:model} evaluates the downstream behavior of the integrated agent model to verify that the closed-loop integration behaves as intended during runtime. Finally, Sec.~\ref{sec:evaluation:scalability} evaluates scalability with respect to runtime performance and memory consumption.

\subsection{Evaluation of the Integration Architecture}
\label{sec:evaluation:architecture}

The primary objective of this section is to evaluate the integration architecture independently of the integrated agent model. In contrast to tool-specific integrations, the architecture follows a modular design based on established formats and is intended to be reusable across different simulation tools. To evaluate the architecture systematically, we apply the evaluation criteria introduced earlier: standards compliance, portability and reusability, integration effort, and result consistency. These criteria reflect our design goals and align with established principles for model integration in simulation-based testing.

\begin{itemize}
  \item \textbf{Standards Compliance:}  
  Our architecture relies exclusively on established standards: \gls{osi} for sensor, and object-level communication and \gls{fmi} for modular model encapsulation. No custom extensions or tool-specific bindings are required. This guarantees long-term interoperability and aligns with standardization efforts in the automated driving domain.
  
  \item \textbf{Portability and Reusability:}  
  The same \gls{osmp}-packaged model can be used without modification in all simulators evaluated in this work, OpenPASS, CARLA, and CarMaker. Tool-specific scenario definitions and parameter configurations remain unchanged when coupling the model.

  \item \textbf{Integration Effort:}  
  Once a simulation environment provides basic \gls{osi} and \gls{fmi} support, models can be integrated without any additional development effort. No adapter code or simulator-specific logic is required, allowing for streamlined toolchain setups. The same integration approach can also be used for other model types, as long as the required OSI messages are provided by the simulation environment.
  
  \item \textbf{Result Consistency:}  
  Basic model behaviors such as velocity profiles, \gls{thw}, and lane change execution can be evaluated consistently within the defined test scenarios, as discussed in Sec.~\ref{sec:evaluation:model}. However, simulator-internal time stepping and dynamics may differ and therefore affect the exact model results. Accordingly, cross-simulator comparisons should be interpreted with caution.
\end{itemize}

These findings demonstrate that the proposed architecture fulfills the intended integration goals. To further contextualize its advantages and limitations, Table~\ref{tab:architecture-comparison} contrasts our approach with a conventional tool-specific integration strategy. The table also includes the primary intended use case of each approach, further clarifying the scope and limitations of conventional tool-specific integrations in comparison to our standardized architecture.

\renewcommand{\arraystretch}{1.5}
\begin{table}[ht]
\centering
\caption{Proposed \gls{osmp}-based architecture compared to a tool-specific baseline integration}
\label{tab:architecture-comparison}
\begin{tabularx}{\linewidth}{
  >{\raggedright\arraybackslash}p{0.2\linewidth} 
  >{\raggedright\arraybackslash}X 
  >{\raggedright\arraybackslash}X
}
\toprule
\textbf{Criterion} &
\textbf{Proposed \gls{osmp}-based Architecture} &
\textbf{Tool-Specific Integration} \\
\midrule
{Intended Use Case} &
Cross-platform integration of simulation models in simulators for ADAS/AD &
Direct integration of custom models into a specific simulation tool \\
{Standards Compliance} &
Fully based on ASAM \gls{osi} and \gls{fmi} standards; open and extensible &
Often proprietary; tight coupling to internal APIs \\
{Portability \& Reusability} &
Same \gls{fmu} and parameters used across tools; no modifications needed &
Model tightly coupled to one simulator; not portable without redevelopment \\
{Model Integration} &
Minimal effort if model supports \gls{osmp} and simulator provides required \gls{osi} data &
Depends on model complexity \\
{Result Consistency} &
depends on simulator core; integration code is identical, which improves consistency &
depends on simulator and specific integration; results can vary due to implementation details \\
{Limitations} &
Initial \gls{osmp} support required; simulator internals only accessible via \gls{osi}/\gls{fmi} &
Tightly coupled to specific tool; inconsistent behavior across platforms \\
\bottomrule
\end{tabularx}
\end{table}
\renewcommand{\arraystretch}{1.0}

As shown in Table~\ref{tab:architecture-comparison}, the \gls{osmp}-based architecture offers advantages in portability, standard compliance, and maintainability compared to conventional tool-specific integrations. It requires initial \gls{osmp} support in the simulation toolchain, but enables modular and reproducible simulation workflows once available.

To complement the architecture-focused discussion, the following subsection assesses whether the reference implementation yields plausible results for the agent model in selected scenarios.

\subsection{Evaluation of Agent Model Results}
\label{sec:evaluation:model}

We demonstrate the closed-loop behavior of the integrated agent model in a set of representative scenarios and inspect the resulting outputs with respect to behavioral plausibility and consistency within the test setup. All scenarios are executed in OpenPASS, while similar qualitative results were observed using the other considered simulation tools.

\subsubsection*{Scenario 1: Following}

\begin{figure}[htbp]
  \centering  \input{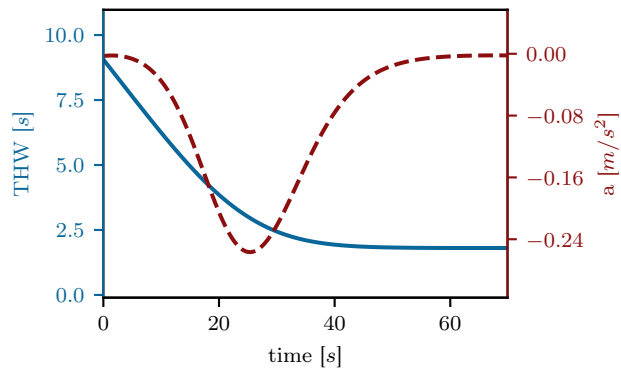}
  \caption{Following scenario: THW (solid plot) and acceleration (dashed plot) of the agent over time.}
  \label{fig:following-plot}
\end{figure}

In the first scenario, the agent starts in free driving and then approaches a slower vehicle ahead. The \gls{thw} and resulting acceleration are visualized in Fig.~\ref{fig:following-plot}. The \gls{thw} decreases continuously and converges toward a stable, parametrizable value of \SI{2}{\second}. The predictive dynamics controller applies smooth deceleration with a minimum of approximately \SI{-0.25}{\metre\per\second\squared}.

\subsubsection*{Scenario 2: Speed Adaptation}
In the second scenario, the agent starts at \SI{70}{\kilo\metre\per\hour} and approaches a \SI{30}{\kilo\metre\per\hour} speed limit sign. As described in Sec.~\ref{sec:implementation:agent-model:behavior-model}, the model reduces its speed predictively and reaches the desired velocity before passing the sign. 

\begin{figure}[h]
  \centering
  \input{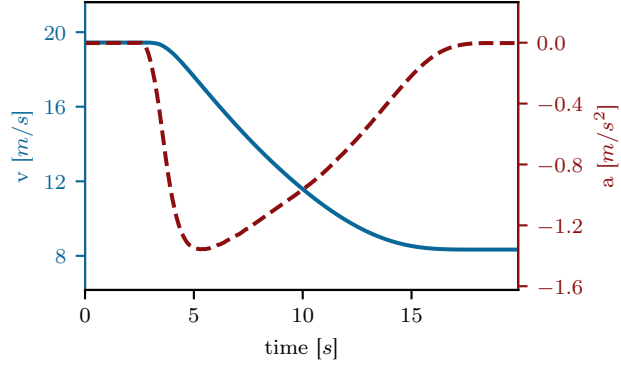}
  \caption{Speed adaptation scenario: Predictive velocity (solid plot) and desired acceleration (dashed plot) of the agent over time.}
  \label{fig:signal-plot}
\end{figure}

Fig.~\ref{fig:signal-plot} shows a smooth velocity reduction that starts about \SI{10}{\second} before the speed sign. This preview time is parameterized via a dedicated \gls{thw} parameter. The desired acceleration exhibits a smooth deceleration profile.

\subsubsection*{Scenario 3: Lane Change}

\begin{figure}[htbp]
  \centering
  \input{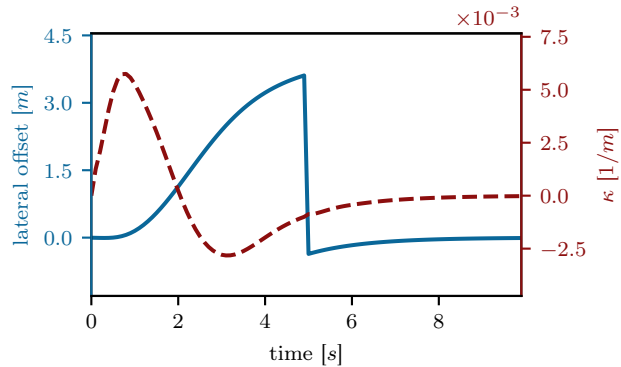}
  \caption{Lane change scenario: lateral offset (solid plot) and curvature (dashed plot) of the agent over time.}
  \label{fig:lanechange-plot}
\end{figure}

In the third scenario, the lateral component of the agent model is investigated. The objective is a route-based lane change guided by a lateral offset from the lane centerline. As shown in Fig.~\ref{fig:lanechange-plot}, the lane change is completed within \SI{5}{\second} and transitions smoothly to the new lane's centerline at a lateral distance of \SI{4}{\metre}. The desired curvature is calculated based on dynamically shifted reference points.

Fig.~\ref{fig:lanechange-plot} indicates a starting positive curvature to initiate the lane change, following a negative curvature to adjust to the new lane. The resulting curvature shows a smooth transition without oscillations. The apparent discontinuity in the lateral offset is caused by reassigning the reference from the current lane to the target lane.

\subsubsection*{Scenario 4: Predictive Speed}

\begin{figure}[h]
  \centering
  \input{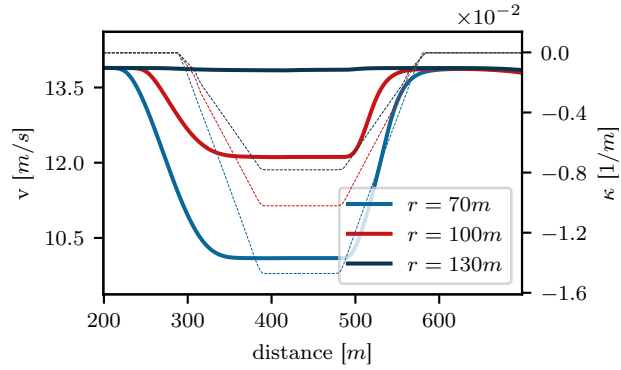}
  \caption{Predictive velocity (solid plots) adaptation based on different road curvatures (dashed plots). The desired agent velocity of \SI{13.88}{\metre\per\second} is reduced depending on the curvature.}
  \label{fig:predictive-speed-plot}
\end{figure}

An additional capability of the agent model is the predictive speed regulation based on the road layout ahead, as already described in Sec.~\ref{sec:implementation:agent-model:behavior-model}. To test predictive speed control, a parameterizable road consisting of a line-spiral-arc-spiral-line layout with variable curvature is created within an OpenDRIVE road generation tool~\cite{Becker2020}. The curvature profiles for three road layouts with a radius of \SI{70}{\metre}, \SI{100}{\metre}, and \SI{130}{\metre} are shown schematically by the dashed lines in Fig.~\ref{fig:predictive-speed-plot}.

Additionally, the agent's velocity along the road distance can be determined for a maximal desired velocity of \SI{50}{\kilo\metre\per\hour} and every curvature layout, respectively. The tighter the turn, the lower the speed in the curve, which is dependent on a maximum comfortable lateral acceleration. It is noteworthy that the speed is reduced even before reaching the curve and increased on the outward spiral. For an arc radius of \SI{130}{\metre}, the desired speed stays constant, which can be explained by the lower calculated limiting curve radius of \SI{128}{\metre} for a desired velocity of \SI{50}{\kilo\metre\per\hour}. Thus, the speed is only reduced for smaller radii.

\subsection{Evaluation of Scalability}
\label{sec:evaluation:scalability}
\begin{figure}[h]
  \centering
  \includegraphics[width=0.8\linewidth]{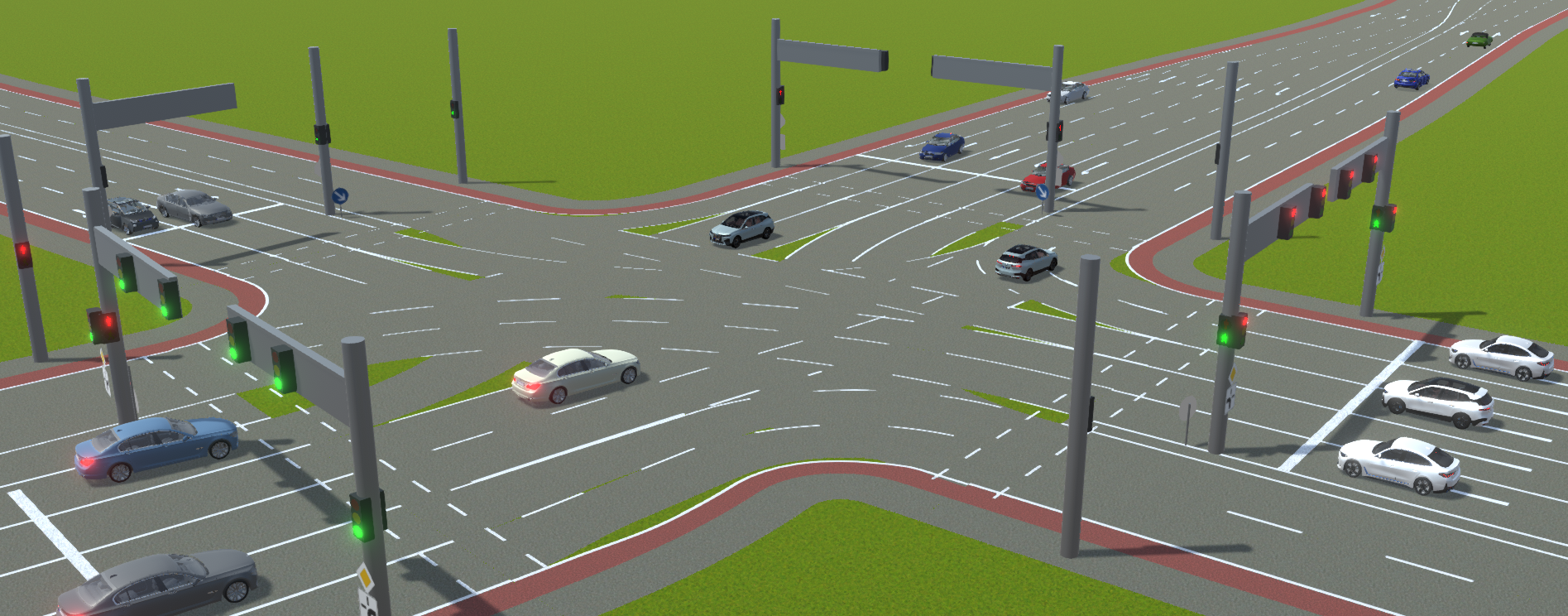}
  \caption{Visualization of the multi-agent interaction scenario executed in OpenPASS. All 20 agents are controlled in parallel with a dedicated agent model instance. The agents comply with traffic rules and signals, avoid collisions, and all reach their target destinations.}
  \label{fig:multi-agent-scenario}
\end{figure}

To assess both functional robustness and performance scalability, we construct a full-scale integration scenario in OpenPASS involving a complex intersection, traffic lights, and dense multi-agent interactions. In the baseline configuration, 20 agents are simulated in parallel, each controlled by an individual instance of the proposed agent model. The scenario has a duration of \SI{60}{\second}. All agents successfully reach their target positions without collisions or traffic rule violations, confirming stable closed-loop behavior under multi-agent conditions. A visualization of this scenario is shown in Fig.~\ref{fig:multi-agent-scenario}.

\begin{figure}[h]
  \centering
  \input{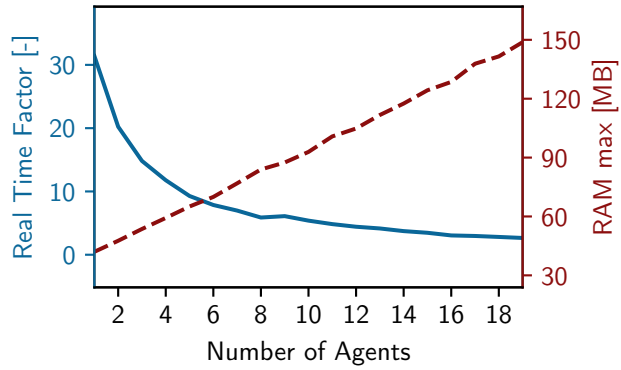}
  \caption{Scalability evaluation of the OpenPASS agent model integration: real time factor (solid plot) and memory consumption (dashed plot) depending on the simulated agent number.}
  \label{fig:scalability}
\end{figure}

Building on this validated 20-agent setup, we evaluate the scalability of the reference implementation with respect to runtime performance and memory consumption. The number of concurrently simulated agents is varied while measuring the total simulation runtime and the peak memory usage of the OpenPASS process. The results in Fig.~\ref{fig:scalability} show an approximately linear increase in both runtime and memory consumption as the number of agents grows. This is expected, as each additional agent requires \gls{osi} processing in OpenPASS and individual model execution. Since the integration does not employ multithreading, one CPU core remains close to full utilization throughout the experiments.

Although memory usage increases linearly with the number of agents, the per-agent footprint remains moderate at approximately \SI{7.5}{\mega\byte} per instance. The measurements further indicate a high real-time factor for small agent numbers, which is partly attributable to the lightweight nature of OpenPASS. For example, with a single agent, the full scenario can be completed in about \SI{2}{\second}, corresponding to a real-time factor of roughly 30.

Overall, the results demonstrate that the integration architecture maintains stable closed-loop behavior while scaling approximately linearly in computational cost, making it suitable for large-scale multi-agent experiments.

\section{Conclusion} 
\label{sec:conclusion}

This article presented a standardized and modular integration architecture for closed-loop simulation models, with a focus on traffic agent behavior. The approach builds upon established standards, specifically \gls{osi} for structured data exchange and \gls{fmi} for encapsulated model interfaces. The \gls{osmp} framework enables seamless model coupling across heterogeneous simulation tools, eliminating the need for tool-specific adaptations.

To demonstrate applicability, we integrated a representative agent model into three different simulation tools: OpenPASS, CARLA, and CarMaker. Basic driving scenarios were used to validate consistent and plausible agent behavior, showcasing the architecture's portability and runtime stability. In addition, a scalability evaluation indicates an approximately linear increase in runtime and memory consumption with the number of simulated agents, supporting applicability to larger multi-agent experiments. Additionally, an automatic end-to-end testing framework with CI integration was developed to support continuous validity against a predefined test catalog.

Our reference implementation including the agent model and test scenarios are made publicly available to foster reproducible research and accelerate the harmonization of interoperable simulation tools. By offering this reference implementation, this work supports the broader adoption of \gls{osi} and \gls{fmi} within simulation tools and facilitates scalable, tool-independent integration across multiple simulation-driven testing frameworks.

\begin{funding} The work of this manuscript has been done in the context of the Horizon Europe project SUNRISE. These projects have been funded by the European Union under grant agreement No. 101069573. Views and opinions expressed are those of the author(s) only and do not necessarily reflect those of the European Union or the European Climate, Infrastructure and Environment Executive Agency (CINEA). Neither the European Union nor the granting authority can be held responsible for them.

Additionally, this work received funding from the SET Level and VVM projects as part of the PEGASUS project family, promoted by the German Federal Ministry for Economic Affairs and Energy based on a decision of the Deutsche Bundestag.

\end{funding}

\printbibliography
 
\end{document}